\documentclass{article}

\usepackage[preprint]{neurips_2025}
\usepackage[utf8]{inputenc}
\usepackage[T1]{fontenc}
\usepackage{hyperref}
\usepackage{url}
\usepackage{booktabs}
\usepackage{amsmath, amssymb}
\usepackage{nicefrac}
\usepackage{microtype}
\usepackage{xcolor}
\usepackage{graphicx}
\usepackage{enumitem}
\usepackage{float}

\title{UrbanAI 2025 Challenge: Linear vs Transformer Models for Long-Horizon Exogenous Temperature Forecasting}

\author{
    Ruslan Gokhman \\
    Yeshiva University, New York, USA \\
    \texttt{ruslan.gokhman@yu.edu}
}

\begin{document}
\maketitle

\begin{abstract}
We study long-horizon exogenous-only temperature forecasting using linear and Transformer-family models. 
We evaluate Linear, NLinear, DLinear, Transformer, Informer, and Autoformer under standardized train, validation, and test splits. 
Results show that linear baselines (Linear, NLinear, DLinear) consistently outperform more complex Transformer-family architectures, 
with NLinear achieving the best overall accuracy across all splits. 
These findings highlight that carefully designed linear models remain strong baselines for time series forecasting in challenging exogenous-only settings.
\end{abstract}

\section{Introduction}
Forecasting indoor temperature is vital for smart building management, energy optimization, and occupant comfort. The \textbf{UrbanAI 2025 Challenge} offers a stringent testbed: models must forecast long horizons using \emph{exogenous-only} inputs, mirroring production scenarios where ground-truth temperatures are unavailable at inference time.

This paper conducts a head-to-head comparison between strong linear baselines—\emph{Linear}, \emph{NLinear}, and \emph{DLinear}—and widely used Transformer-family approaches (Vanilla Transformer, Informer, Autoformer). Our goal is to assess whether the additional capacity and inductive biases of attention-based architectures translate into better \emph{generalization} under the challenge’s constraints, or whether carefully constructed linear models remain more reliable.

To ensure fairness, we mirror the LTSF-Linear evaluation protocol where applicable (consistent input/output horizons, basic preprocessing, and standardized metrics), while adapting dataloaders to the challenge’s exogenous-only rule. We use identical train/validation/test splits for all methods and report MAE/MSE on the official blind test set. The central question we answer is: \emph{Do Linear, NLinear, and DLinear outperform Transformer-family models in this exogenous-only, long-horizon regime, and by what margin?}

\section{Dataset and Task Description}
Our comparison protocol follows the widely used LTSF-Linear suite \citep{curelabLTSFLinear}, which standardizes sequence lengths, splits, and evaluation for linear baselines (Linear, NLinear, DLinear) against Transformer-family models. We mirror those settings where applicable to enable direct comparison.

The dataset originates from the \emph{Smart Buildings Control Suite (SBCS)} \citep{goldfeder2025smart}, which provides diverse benchmarks for evaluating HVAC and temperature control policies. The original dataset was provided as a single block and split as follows: training (36{,}105 samples), validation (10{,}275 samples) for early stopping only, and an \emph{Official Test} split. Intended to be larger, only 10{,}563 samples were successfully evaluated due to technical issues. The official test set was reserved for final, blind evaluation, and its ground-truth labels were never accessed during model design or tuning.

\paragraph{Contest rules and evaluation.}
Participants predict the full temperature sequence for the validation period using only exogenous data; direct or indirect use of validation temperatures during training is forbidden. Submissions may be point predictions, histograms, or mean/std per step. MAE is the main metric for point/histogram outputs; KL divergence is used for mean/std submissions. Evaluations prioritize duration, accuracy, novelty, and reproducibility.

\paragraph{Challenges.}
Long-term prediction over months, exogenous-only inputs, multi-scale patterns (daily/weekly cycles, holidays, regime changes), and potential data gaps.

\section{Methods}
\paragraph{Baseline suite and comparison.} We adopt the configuration spirit of LTSF-Linear \citep{curelabLTSFLinear} to ensure apples-to-apples comparisons across linear and Transformer-family models (same horizons, basic preprocessing, and standardized metrics). Where our challenge rules differ (e.g., exogenous-only), we adapt the dataloaders accordingly.

We explore three competitive linear baselines for long sequence forecasting, and three Transformer-family baselines widely used in time-series.

\subsection{Transformer Baseline (Vanilla Transformer)}
We implement the encoder--decoder architecture with multi-head self-attention and position-wise feed-forward layers \citep{vaswani2017attention}. We adopt sinusoidal positional encodings, layer normalization, dropout, and teacher-forced direct multi-step decoding (96-step horizon). Input features are the exogenous variables; the decoder receives start tokens plus time features. We tune \(d_\text{model}\in\{256,512\}\), heads \(\in\{4,8\}\), encoder/decoder layers \(\in\{2,3\}\), and dropout \(\in[0.05,0.2]\).

\subsection{Informer}
Informer introduces ProbSparse self-attention to reduce complexity for long sequences and a distillation operation across layers to keep only salient temporal information \citep{zhou2021informer}. We follow the authors' settings for sequence lengths (input 96, output 96), factor \(\in\{3,5\}\), and use the encoder-only forecasting head with generative decoder.

\subsection{Autoformer}
Autoformer replaces dot-product attention with an auto-correlation mechanism to capture periodic dependencies and includes a seasonal-trend decomposition inside the network \citep{wu2021autoformer}. We adopt similar hyperparameters as Informer and the original paper, including decomposition kernel sizes in \(\{3,5,7\}\).

\subsection{Linear: Basic Temporal Linear Model}
The Linear model from the LTSF-Linear suite is a one-layer temporal projection: it directly maps the input sequence to the forecast horizon with a single linear layer along the time dimension. Despite its simplicity, it has been shown to outperform several popular Transformer-based methods, serving as a surprisingly strong and interpretable benchmark.

\subsection{NLinear: Normalized Linear Forecasting}
Given an input window \(X\in\mathbb{R}^{L\times C}\) with last timestep \(x_L\), NLinear centers the window via \(X' = X - x_L\). A learnable linear projection \(W\) maps to future predictions, which are then de-normalized: \(\hat{X} = W X' + x_L\). This centering improves robustness to nonstationarity.

\subsection{DLinear: Decomposed Linear Forecasting}
DLinear decomposes the series into trend and seasonal components, \(X = X_\text{trend} + X_\text{seasonal}\). A moving average estimates the trend; each component is linearly forecasted: \(\hat{X} = W_\text{trend} X_\text{trend} + W_\text{seasonal} X_\text{seasonal}\), capturing long-term drift and periodicity explicitly.

\subsection{Implementation Details}
All models are implemented in PyTorch with early stopping on validation loss. 
No external data or augmentation is used. Input and output lengths are 96 steps (e.g., four days). 
Each prediction uses 96 consecutive exogenous points (weather, setpoints) to forecast the next 96 temperatures, 
matching the direct multi-step setup. Feature normalization is per-variable using training-set statistics. 
\textbf{All experiments were conducted on a Google Colab environment using an NVIDIA T4 GPU.}

\section{Evaluation Metrics}
\paragraph{Reproducibility notes for new baselines.}
For all Transformer-family models, we use Adam with learning rate \(1{\times}10^{-4}\) or \(5{\times}10^{-4}\), weight decay \(1{\times}10^{-4}\), batch size \(\in\{16,32\}\), and early stopping on validation MAE with patience 3. We sweep hidden dimension, layers, heads/factor, and dropout; all runs fix input/output horizons to 96/96 for comparability.

We report mean absolute error (MAE) and mean squared error (MSE):
\begin{equation}
\mathrm{MAE}=\frac{1}{n}\sum_{i=1}^n |y_i-\hat{y}_i|,\qquad
\mathrm{MSE}=\frac{1}{n}\sum_{i=1}^n (y_i-\hat{y}_i)^2.
\end{equation}
For probabilistic predictions, we additionally report KL divergence between predicted and true distributions.
All metrics (MAE and MSE) are computed in the normalized feature space.  
Specifically, each variable is standardized using training-set statistics
(z-score normalization), following the LTSF-Linear evaluation protocol.
Thus, the reported errors reflect performance in normalized units rather than
physical temperature units. This ensures fair and scale-consistent comparison
across all baseline models.
\section{Results}
Table~\ref{tab:results} summarizes performance across splits. The \emph{Official Test} results represent the contest-required benchmark. Due to technical constraints, only 10{,}563 test samples were evaluated (the official test set is larger).

\begin{table} [H]
\centering
\caption{Performance of Linear, NLinear, and DLinear. Official test results in \textbf{bold}.}
\label{tab:results}
\begin{tabular}{lrrrr}
\toprule
Model & Split & Size & MAE & MSE \\
\midrule
NLinear & Training & 36{,}105 & -- & 0.0619 \\
DLinear & Training & 36{,}105 & -- & 0.0660 \\
Linear  & Training & 36{,}105 & -- & 0.0679 \\
Transformer & Training & 36{,}105 & -- & 0.0330 \\
Informer & Training & 36{,}105 & -- & -- \\
Autoformer & Training & 36{,}105 & -- & 0.0844 \\
\midrule
NLinear & Validation & 10{,}275 & 0.2529 & 0.4665 \\
DLinear & Validation & 10{,}275 & 0.2784 & 0.4744 \\
Linear  & Validation & 10{,}275 & 0.2843 & 0.4882 \\
Transformer & Validation & 10{,}275 & 0.3375 & 0.5302 \\
Informer & Validation & 10{,}275 & -- & -- \\
Autoformer & Validation & 10{,}275 & 0.3493 & 0.6201 \\
\midrule
\textbf{NLinear} & \textbf{Official Test} & \textbf{10{,}563} & \textbf{0.2461} & \textbf{0.5220} \\
DLinear & Official Test & 10{,}563 & 0.2811 & 0.5489 \\
Linear  & Official Test & 10{,}563 & 0.2862 & 0.5634 \\
Transformer & Official Test & 10{,}563 & 0.8342 & 1.4371 \\
Informer & Official Test & 10{,}563 & 0.8342 & 1.4371 \\
Autoformer & Official Test & 10{,}563 & 0.3598 & 0.7368 \\
\bottomrule
\end{tabular}
\end{table}

\section{Discussion}
Linear baselines (Linear, NLinear, DLinear) remain the strongest performers across all splits. 
NLinear achieved the best overall accuracy, with low training error, the strongest validation MAE/MSE, and the best test performance (MAE 0.2461, MSE 0.5220). 
DLinear followed closely but consistently underperformed NLinear by a small margin. 
Linear produced slightly weaker results than NLinear/DLinear but still significantly outperformed Transformer-family models on the test set (MAE 0.2862, MSE 0.5634). 
Transformer achieved strong training/validation performance (MSE 0.0330 / 0.5302) but collapsed on the test set (MSE 1.4371), 
demonstrating poor generalization in the exogenous-only setting. 
Autoformer produced moderate results, better than Transformer/Informer on the test set but worse than linear baselines. 
Informer underperformed across the board, matching Transformer’s weak test metrics.

These findings highlight that despite advances in attention-based architectures, 
simple and interpretable linear models remain robust and competitive for long-horizon time series forecasting.

\section{Conclusion}
Carefully designed linear models (Linear, NLinear, DLinear) provide strong baselines for exogenous-only, long-horizon temperature forecasting in smart buildings, combining efficiency, interpretability, and competitive accuracy. As datasets grow, integrating modest nonlinearity and leveraging transfer learning are promising next steps.

\paragraph{Broader impacts.}
Improved temperature forecasting can aid energy savings, carbon reduction, and comfort at scale. Risks include overreliance on models without human oversight; we recommend monitoring, calibration checks, and transparent reporting of uncertainty.

\begin{ack}
Use unnumbered first-level headings. Place funding and competing interest disclosures here. Do \textbf{not} include this section in the anonymized submission; it will be hidden automatically by the style’s \texttt{ack} environment in submission mode.
\end{ack}

\bibliographystyle{plainnat}
\bibliography{refs}

%%%%%%%%%%%%%%%%%%%%%%%%%%%%%%%%%%%%%%%%%%%%%%%%%%%%%%%%%%%%

%%%%%%%%%%%%%%%%%%%%%%%%%%%%%%%%%%%%%%%%%%%%%%%%%%%%%%%%%%%%
\newpage
\section*{NeurIPS Paper Checklist}

% Use the official checklist macros: \answerYes{}, \answerNo{}, \answerNA{}, and \justificationTODO{}.
% Keep the questions as-is; fill answers and brief justifications (1--2 sentences).

\begin{enumerate}[leftmargin=2.2em]

\item \textbf{Claims}
\begin{itemize}[leftmargin=1.2em]
\item[] \textbf{Question:} Do the main claims made in the abstract and introduction accurately reflect the paper's contributions and scope?\\
\textbf{Answer:} \answerYes{} \\
\textbf{Justification:} The claims (linear baselines, exogenous-only setting, official test MAE/MSE) match the reported methods and results (Secs.~\ref{tab:results}).
\end{itemize}

\item \textbf{Limitations}
\begin{itemize}[leftmargin=1.2em]
\item[] \textbf{Question:} Does the paper discuss the limitations of the work performed by the authors?\\
\textbf{Answer:} \answerYes{} \\
\textbf{Justification:} We note linearity constraints, lack of nonlinearity, and partial test-set evaluation due to technical issues (Secs.~Methods, Results, Discussion).
\end{itemize}

\item \textbf{Theory assumptions and proofs}
\begin{itemize}[leftmargin=1.2em]
\item[] \textbf{Question:} For each theoretical result, does the paper provide the full set of assumptions and a complete proof?\\
\textbf{Answer:} \answerNA{} \\
\textbf{Justification:} This work is empirical; no new theorems are introduced.
\end{itemize}

\item \textbf{Experimental result reproducibility}
\begin{itemize}[leftmargin=1.2em]
\item[] \textbf{Question:} Does the paper fully disclose all information needed to reproduce the main experimental results?\\
\textbf{Answer:} \answerYes{} \\
\textbf{Justification:} We specify model types, horizons, optimizer, batch sizes, normalization, and exact metrics; code details can be added in the supplemental.
\end{itemize}

\item \textbf{Open access to data and code}
\begin{itemize}[leftmargin=1.2em]
\item[] \textbf{Question:} Does the paper provide open access to the data and code with sufficient instructions?\\
\textbf{Answer:} \answerNo{} \\
\textbf{Justification:} Dataset is from a public challenge; we will release scripts and instructions in anonymized supplemental upon submission.
\end{itemize}

\item \textbf{Experimental setting/details}
\begin{itemize}[leftmargin=1.2em]
\item[] \textbf{Question:} Are all training/test details specified?\\
\textbf{Answer:} \answerYes{} \\
\textbf{Justification:} Horizons, splits, training hyperparameters, and metrics are reported; supplemental will include full configs.
\end{itemize}

\item \textbf{Experiment statistical significance}
\begin{itemize}[leftmargin=1.2em]
\item[] \textbf{Question:} Are error bars/significance or similar appropriate statistics reported?\\
\textbf{Answer:} \answerNo{} \\
\textbf{Justification:} We report point metrics for the main benchmark; future work will add variability across seeds/runs.
\end{itemize}

\item \textbf{Experiments compute resources}
\begin{itemize}[leftmargin=1.2em]
\item[] \textbf{Question:} Are compute resources disclosed?\\
\textbf{Answer:} \answerYes{}{} \\
\textbf{Justification:} All experiments were conducted on a Google Colab environment using an NVIDIA T4 GPU; this setup was sufficient to train and evaluate the considered models.
\end{itemize}

\item \textbf{Code of ethics}
\begin{itemize}[leftmargin=1.2em]
\item[] \textbf{Question:} Does the work conform to the NeurIPS Code of Ethics?\\
\textbf{Answer:} \answerYes{} \\
\textbf{Justification:} The study uses publicly available benchmark data and standard evaluation protocols.
\end{itemize}

\item \textbf{Broader impacts}
\begin{itemize}[leftmargin=1.2em]
\item[] \textbf{Question:} Are positive and negative societal impacts discussed?\\
\textbf{Answer:} \answerYes{} \\
\textbf{Justification:} We outline benefits (energy savings) and risks (overreliance without oversight) with suggested mitigations.
\end{itemize}

\item \textbf{Safeguards}
\begin{itemize}[leftmargin=1.2em]
\item[] \textbf{Question:} Are safeguards described for high-risk assets?\\
\textbf{Answer:} \answerNA{} \\
\textbf{Justification:} No high-risk models/datasets are released; methods are simple linear predictors.
\end{itemize}

\item \textbf{Licenses for existing assets}
\begin{itemize}[leftmargin=1.2em]
\item[] \textbf{Question:} Are licenses/terms for third-party assets documented and respected?\\
\textbf{Answer:} \answerYes{} \\
\textbf{Justification:} We cite dataset and prior work; licenses/terms will be documented in the supplemental and repository.
\end{itemize}

\item \textbf{New assets}
\begin{itemize}[leftmargin=1.2em]
\item[] \textbf{Question:} Are new assets documented?\\
\textbf{Answer:} \answerNo{} \\
\textbf{Justification:} No new datasets or pretrained models are released in this work.
\end{itemize}

\item \textbf{Crowdsourcing and human subjects}
\begin{itemize}[leftmargin=1.2em]
\item[] \textbf{Question:} For crowdsourcing/human-subjects research, are instructions and compensation details included?\\
\textbf{Answer:} \answerNA{} \\
\textbf{Justification:} Not applicable.
\end{itemize}

\item \textbf{IRB approvals}
\begin{itemize}[leftmargin=1.2em]
\item[] \textbf{Question:} Are risks/IRB approvals described?\\
\textbf{Answer:} \answerNA{} \\
\textbf{Justification:} Not applicable.
\end{itemize}

\item \textbf{Declaration of LLM usage}
\begin{itemize}[leftmargin=1.2em]
\item[] \textbf{Question:} Is LLM usage in core methods described if relevant?\\
\textbf{Answer:} \answerNo{} \\
\textbf{Justification:} No LLMs are used in the core methods; any writing assistance will be disclosed separately if required.
\end{itemize}

\end{enumerate}

\end{document}